\documentclass[letterpaper, 10 pt, conference]{ieeeconf}  

\IEEEoverridecommandlockouts                              

\usepackage{graphics} 
\usepackage{graphicx}
\usepackage{subfigure}
\usepackage{multirow}
\usepackage{booktabs}
\usepackage{amsmath}
\usepackage{url}
\usepackage{mathrsfs}

\title{\LARGE \bf
Semantic Flow-guided Motion Removal Method for Robust Mapping
}

\author{Xudong Lv$^{*}$, Boya Wang$^{*}$, Dong Ye, and Shuo Wang
\thanks{Corresponding author: Shuo Wang}
\thanks{* contributed equally to this work}
\thanks{Xudong Lv, Boya Wang, Dong Ye, and Shuo Wang are with School of Instrumentation Science and Engineering, Harbin Institute of Technology, Harbin 150001, China (email: 15B901019@hit.edu.cn; 19B901034@stu.hit.edu.cn; yedong@hit.edu.cn; 15B901018@hit.edu.cn)}}

\begin{document}

\maketitle
\thispagestyle{empty}
\pagestyle{empty}

\begin{abstract}

Moving objects in scenes are still a severe challenge for the SLAM system. Many efforts have tried to remove the motion regions in the images by detecting moving objects. In this way, the keypoints belonging to motion regions will be ignored in the later calculations.  In this paper, we proposed a novel motion removal method, leveraging semantic information and optical flow to extract motion regions. Different from previous works, we don’t predict moving objects or motion regions directly from image sequences. We computed rigid optical flow, synthesized by the depth and pose, and compared it against the estimated optical flow to obtain initial motion regions. Then, We utilized K-means to finetune the motion region masks with instance segmentation masks. The ORB-SLAM2 integrated with the proposed motion removal method achieved the best performance in both indoor and outdoor dynamic environments. 

\end{abstract}

\section{INTRODUCTION}
Simultaneous Localization and Mapping (SLAM) is a key technology in robotics, automation, and computer vision. Most SLAM systems work well in static environments but fail in dynamic environments (such as crowded shopping malls). Algorithms tend to fail in the presence of dynamic objects due to errors in frame tracking, loop closure, and local mapping. Therefore, how SLAM system building maps in dynamic environments is a research focus nowadays. Inertial Measurement Unit (IMU) and Random Sample Consensus(RANSAC) can deal with slightly dynamic environments to some extent. Motion removal is a more direct and better way to eliminate the impact of moving objects on SLAM. Classical motion removal methods, such as background subtraction mentioned in \cite{Dhome2010A} and interframe differences mentioned in \cite{zhencha}, are based on the assumption that the camera is fixed. Once the camera angle or position changes, these methods will be invalid. \cite{Sheikh2009Background} are proposed to deal with the problem of a moving camera in the dynamic scene. \cite{Kitt2010Moving} combine ego-motion estimation and low-level object detection to remove moving objects. RANSAC has been used for motion removal in \cite{Litomisky2012Removing} and \cite{7418963}. 

\begin{figure}[thpb]
    \centering
    \includegraphics[width=8cm]{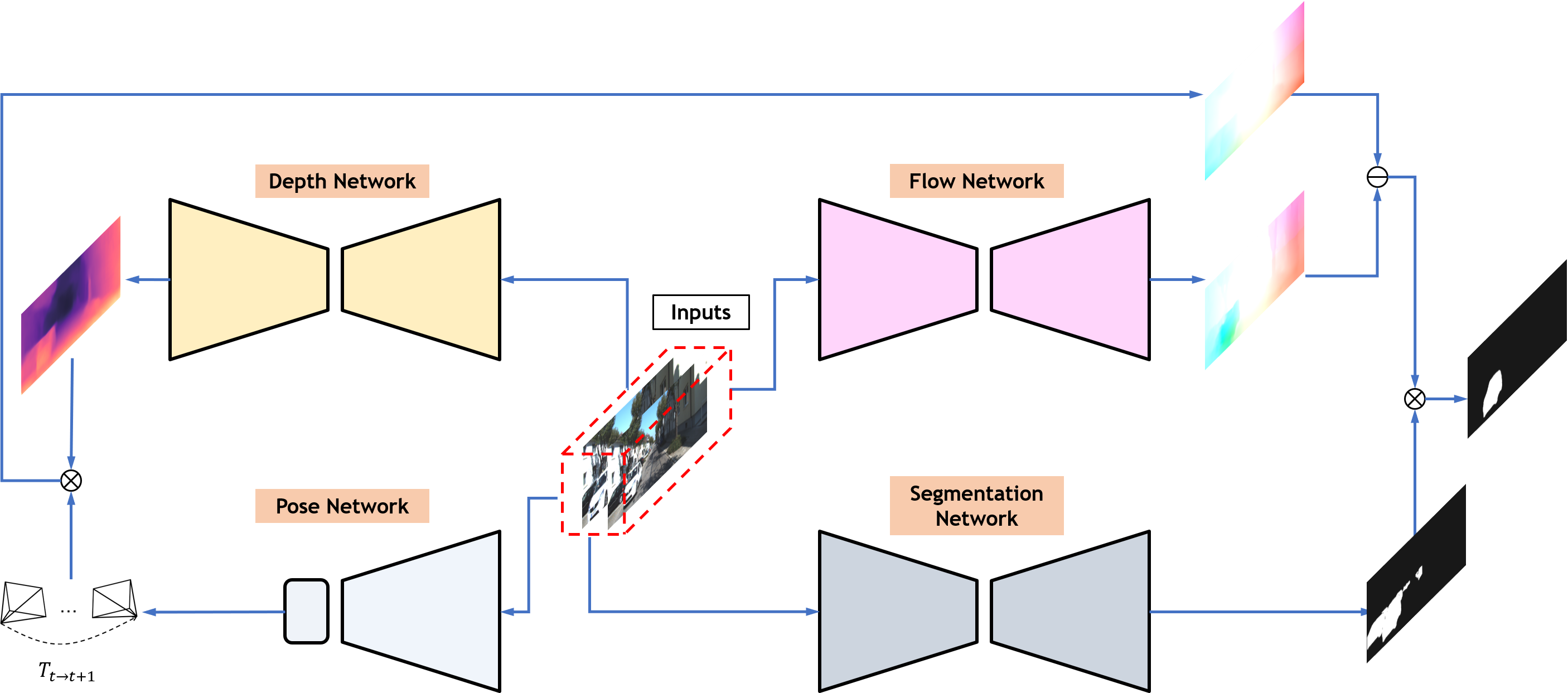}
    \caption{The workflow of our proposed Semantic Flow-guided motion removal method. In our workflow, two adjacent frames are taken as input to four CNNs. Depth Network generates a depth map for ${{I}_{t}}$, while Pose Network outputs the camera pose from ${{I}_{t}}$ to ${{I}_{t\text{+1}}}$. Rigid flow from ${{I}_{t}}$ to ${{I}_{t\text{+1}}}$ can be synthesized by the corresponding depth map and the camera pose. Then, we can get an initial motion mask by rigid flow and optical flow predicted through Flow Network. We use a Segmentation Network to predict an instance semantic mask for ${{I}_{t}}$, and the instance mask can be adopted to finetune the initial motion mask. Finally, we get a semantic flow-guided mask to remove moving objects from ${{I}_{t}}$.}
    \label{workflow}
\end{figure}

Object detection or semantic segmentation based on deep learning can detect objects and determine objects' motion state according to their category attributes. However, such approaches are rough. Some detected objects may be static at that time but still will be removed based on category. For example, in some parking lots or some sequences of KITTI data, many vehicles are static. If feature points on these static objects are ignored, the accuracy and stability of localization and mapping will decrease, or even mapping will fail. Therefore, in addition to the category of objects, motion information should be considered. Bayesian can be used to update the motion state of feature points. But a priori knowledge of the motion of objects is hard to obtain, significantly when the scene is continually changing. Therefore, in the absence of prior knowledge, how to accurately get the motion state of objects in the dynamic environments is still a severe challenge.

Optical flow is another way to estimate the motion state of objects. The motion information can be obtained using the change of pixels in the time domain and the correlation between adjacent frames. The moving objects can be judged directly from the optical flow diagram when the camera is static. When the camera moves, the moving object can also be detected by combining the computed optical flow diagram and the multi-view geometric constraints based on filtering or optimization. The advantage of the optical flow method is that it can accurately detect and identify the position and the motion state of the moving objects without knowing the scene information. It is still applicable when the camera is moving. The optical flow carries not only motion information of moving objects but also rich information about the three-dimensional structure of the scene.

In this paper, we propose a motion removal method guided by semantic and optical flow. Learning-based methods predict the instance segmentation mask, depth, optical flow, and pose of each frame in parallel. Depth and camera pose can synthesize the rigid flow of the current frame. By utilizing the residual information between the optical flow and the rigid flow, the region of moving objects in the current frame can be obtained. The instance masks are used to finetune the motion regions obtained through the optical flow. In this way, it is possible to distinguish moving objects from the current scene, thus preserving static objects' feature points. This method avoids removing moving objects only based on category attributes, which leads to improper removal of feature points of stationary objects (such as parking cars) and the SLAM system's tracking failure. The contributions of this paper are as follows:
\begin{enumerate}
    \item We propose a method for motion removal guided by semantic and optical flow. The depth and pose of the current frame are used to synthesize the current frame's rigid flow. We utilized the predicted optical flow and the synthesized rigid flow to obtain the motion regions.
    \item In this work, a learning-based instance segmentation network is used to predict the instance semantic mask of potential moving objects in the current frame. The semantic masks are employed to finetune the region of moving objects calculated by optical flow, which is not accurate and smooth enough.
    \item We add our proposed motion removal method to the front-end of the state-of-the-art (SOTA) ORB-SLAM2 \cite{ORBSLAM2AO} system to evaluate the performance. By eliminating the moving objects' influence, the accuracy of localization and mapping of the ORB-SLAM2 is improved in the dynamic scene.
    \item The method proposed in this paper is tested on the KITTI dataset. The experimental results show that the method can be well adapted to the dynamic scene containing many static and movable objects. Besides, the technique is also tested on the TUM dataset. The experimental results show that the performance of our proposed method is better than other motion removal methods.
\end{enumerate}

\section{RELATED WORK}
\subsection{Detection-based Motion Removal}
Deep learning-based object detection or segmentation models are introduced to detect moving objects and remove the motion regions. Detect-SLAM \cite{Detectslam} incorporated Single Shot MultiBox Detector (SSD) \cite{Liu2016SSD} into the ORB-SLAM2 \cite{ORBSLAM2AO}. Moving objects can be removed during the per-frame tracking process, feature points associated with the moving objects were filtered out, and the static objects detected in keyframes are rebuilt. A missing detection compensation algorithm based on velocity invariance and Bayesian scheme was proposed to improve the SSD detector's accuracy. Dynamic-SLAM \cite{2019Dynamic} integrated the SSD object detector to the SLAM system. A missed detection compensation algorithm based on the speed invariance and Bayes' rule is proposed to improve the SSD detector's recall. SOF-SLAM \cite{2019SOF} presented a semantic optical flow SLAM system toward dynamic environments. With the combination of semantic and geometry information, the system can overcome the drawback of sole utilization. DynaSLAM \cite{DynaSLAM} added dynamic target detection and background repair technologies on the base of ORB-SLAM2. The method based on multi-view geometry and Mask RCNN \cite{maskrcnn} was comprehensively used to detect the dynamic objects and obtained a static map of the scene. Then, on this basis, the input frame background was repaired to fill in the area hidden by the dynamic objects. \cite{Nikolas2018Semantic} proposed a probabilistic model to extract the relative static objects to deal with the dynamic environment. The map's inlier ratio was updated by introducing new observation data constantly to realize the smooth transition between dynamic and static map points. DS-SLAM \cite{Qiao2018DS} combined a semantic segmentation network with a motion consistency check method to reduce dynamic objects' influence. DS-SLAM maintained high accuracy in a highly dynamic environment. 

\cite{Coarse} divided objects in the scene into non-static and static motion states through the deep neural network. On this basis, the semantic data association method for motion removal inframe tracking thread and local mapping thread was proposed. The improved CenterNet was introduced as a moving object detection thread to provide semantic information and coarse location. In the improved SLAM system, the authors proposed a novel semantic data association method to project the semantic information from 2D to 3D. According to the semantic categories, moving objects' feature points will not be calculated in the other threads.

\subsection{Optical flow-based motion removal}
Traditional optical flow-based motion removal works in \cite{Namdev2012Motion} \cite{Lim2012Modeling} \cite{detect}  \cite{Zamalieva2014Exploiting}  \cite{6907020} utilized epipolar constraint to eliminate moving objects. Leveraging deep learning, \cite{Fragkiadaki2014Learning} adopted the optical flow method to obtain the optical flow of images and regarded the optical flow images and RGB images as the convolutional neural network's input. The outputs of the convolutional neural network were defined as Moving Objectness. Then the spatiotemporal segment proposals are sorted according to Moving Objectness to divide the image into the moving foreground and static background. \cite{lv2018learning} employed two separate networks to process the input RGB-D images independently. The camera pose and rigid/non-rigid mask were estimated by Rigidity Transform Network (RTN). The dense optical flow was acquired by PWCNet \cite{sun2018pwc}. The predicted camera pose was further refined by the predicted dense flow over the rigid region. FlowFusion \cite{2020arXiv200305102Z} combined deep learning-based optical flow estimation with the RGB-D SLAM system. This algorithm adopted optical flow residuals to highlight the dynamic semantics in the RGB-D point cloud. FlowFusion provided more accurate and efficient rigid/non-rigid segmentation for camera tracking and background reconstruction.

\section{METHOD}

In this section, we describe how to combine semantic and flow information to detect moving objects in a dynamic environment. By eliminating the influence of detected moving objects on frame tracking and local mapping, the accuracy and stability of the localization and mapping can be improved. The framework of our proposed motion removal method is shown in Fig. \ref{workflow}. The framework is mainly composed of two parts: the semantic-guided motion detection module and the flow-guided motion detection module. In the semantic-guided module, a pre-trained instance segmentation network is applied to detect potential moving objects. We adopt three separate networks in the flow-guided module to predict the current frame's depth, optical flow, and camera pose between the adjacent frames. According to the multiple view geometry, the predicted camera pose and depth image can synthesize the rigid flow induced by camera motion. Combining the full optical flow of adjacent frames, we can obtain the flow-guided mask of moving objects. With the appropriate coupling of the potential moving objects and motion mask mentioned above, we can get the finer full mask of the moving objects, namely a semantic-flow guided mask. To verify our proposed algorithm's validity, we add it to the front-end of the SOTA ORB-SLAM2 system.

\subsection{Prediction of the semantic-guided motion mask}
The semantic information can provide prior knowledge for the moving objects, which is beneficial to detect the dynamic objects. Object detection can provide the category and the location of each object in the image. While semantic segmentation can predict the labels for each pixel in the image. Based on object detection and semantic segmentation, different instances of the same category can be distinguished. In this paper, we adopt an instance segmentation network ${{\mathcal{N}}_{instance}}$ to acquire the objects that are potentially dynamic or movable. The potentially dynamic class in this paper is ${C}=\{car, bus, motorcycle, bicycle, truck, person, rider\}$. The output of the network ${{\mathcal{N}}_{instance}}$, assuming that the input is an RGB image ${{I}_{t}}$, is the semantic segmentation mask ${{S}_{t}}$ and the instance segmentation mask $S_{t}^{ins}$ corresponding to ${{I}_{t}}$. The semantic-guided mask $M_{t}^{s}$ for potentially dynamic objects can be acquired by
\begin{equation}
    M_{t}^{s}\text{(}{{p}_{i}}\text{)=}\Pi \text{(}{{S}_{t}}({{p}_{i}})\in {C} \bigodot S_{t}^{ins}({{p}_{i}})
\end{equation}
where $\prod$ is an indicator function and equals 1 if the condition is true, ${{p}_{i}}\in {R}^{2},i=1,2,\ldots ,{{N}_{I}}$ is a pixel of the image ${{I}_{t}}$, ${{N}_{I}}$ is the total number of pixels in the image. $M_{t}^{s}({{p}_{i}})\in \{0,1,2,\ldots ,n\}$, where n is the number of detected objects. $\bigodot$ means pixel-wise product.

\subsection{Prediction of the flow-guided motion mask}
Images or videos are obtained by projecting the 3D space points within the camera’s field of view from the world coordinate to the image coordinate. In practice, the real scene is composed of static background and moving objects. The movement of the static background is caused by the motion of the camera, while the movement of the moving object is determined by both camera motion and object motion. To eliminate the influence of moving objects and improve the accuracy of camera pose estimation, we need to segment the region of the moving objects from the whole image. In this paper, we define the image obtained by projecting the movement of the static background into the camera coordinate as the rigid flow ${{f}^{rig}}$. Given adjacent frames ${{I}_{t}}$ and ${{I}_{t\text{+1}}}$,  the depth prediction network ${{\mathcal{N}}_{depth}}$ and the camera pose prediction network ${{\mathcal{N}}_{pose}}$ are applied to predict the depth map ${{D}_{t}}$ for the frame ${{I}_{t}}$  and the relative camera motion ${{T}_{t\to t+1}}$ from ${{I}_{t}}$ to ${{I}_{t\text{+1}}}$. The relative 2D rigid flow $f_{t\to t+1}^{rig}$ from the current frame ${{I}_{t}}$ to the adjacent frame ${{I}_{t\text{+1}}}$ can be denoted by
\begin{equation}
    f_{t\to t+1}^{rig}\text{(}{{p}_{i}}\text{)=}K{{T}_{t\to t+1}}{{D}_{t}}({{p}_{i}}){{K}^{-1}}{{p}_{i}}-{{p}_{i}}
\end{equation}
where $K$ represents the camera intrinsic and ${{p}_{i}}$ represents homogeneous coordinates of pixels in the frame ${{I}_{t}}$. For the notation brevity of equation 2, we omit the necessary conversion to homogeneous coordinates here similar to \cite{2018GeoNet}.
Unlike the rigid flow $f_{t\to t+1}^{rig}$, the output $f_{t\to t+1}^{full}$ of the optical flow prediction network ${{\mathcal{N}}_{flow}}$ is a complete flow containing the movement of the camera and the specific moving objects. The flow-guided motion mask of the image ${{I}_{t}}$ can be represented by
\begin{equation}
M_{t\to t+1}^{f}=1-\prod (\frac{\left\| f_{t\to t+1}^{full}-f_{t\to t+1}^{rig} \right\|}{\max (\left\| f_{t\to t+1}^{full}-f_{t\to t+1}^{rig} \right\|)}<{{m}_{th}})
\end{equation}
where $\prod $ is an indicator function and equals 1 if the condition is true, $\left\| \cdot  \right\|$ is the L2 normalization, ${{m}_{th}}$ is the motion threshold. The area $M_{t\to t+1}^{f}=0$ belongs to the static background and $M_{t\to t+1}^{f}=0$ belongs to the dynamic objects.  To set the value of ${{m}_{th}}$ adaptively according to a different scene, we utilize the clustering algorithm K-means to acquire the cluster center of the static and dynamic area. According to these two cluster centers, the flow-guided motion mask is denoted by
\begin{equation}
    M_{t\to t+1}^{f}({{p}_{i}})=\begin{cases}
    0&\text{$\Psi (f_{t\to t+1}^{non}({{p}_{i}})={c}_{s}$},\\
    1&\text{$\Psi (f_{t\to t+1}^{non}({{p}_{i}})={c}_{m}$}.
    \end{cases}
\end{equation}
where $\Psi (f_{t\to t+1}^{non}({{p}_{i}}))$ means which cluster center the pixel ${{p}_{i}}$ belongs to, ${{c}_{m}}$ and ${{c}_{s}}$are the cluster centers of the moving area and the static area respectively. 

\subsection{Semantic flow-guided motion mask}
The potential moving objects predicted by the semantic-guided module only contain the category of the objects. Thus, the motion state of the objects cannot be judged. Restricted by the prediction accuracy of the networks, the motion mask predicted by the flow-guided module cannot cover the whole moving objects like the instance segmentation. Some wrong predictions may occur in the area that belongs to a stationary scene, buildings, and vegetations for instance. We can find the limitations in both modules, hence the motivation for their combined use. Benefit from the instance segmentation can accurately detect the category and location of each instance, the motion mask can be regarded as the combination of the motion state of each detected objects. This combination mask is defined as the semantic flow-guided motion mask $M_{t\to t+1}^{sfg}$. We calculate the ratio of the number of pixels belongs to the intersection of each instance mask and motion mask to the number of pixels in the instance mask. If the ratio is greater than a threshold, the instance is considered to be dynamic, otherwise, it is static. Consider an instance object $q\in \{1,2,\ldots ,n\}$ with spatial pixel domain ${{\Omega }_{q}}$, $M_{t\to t+1}^{sfg}$ is represented by
\begin{equation}
    M_{t\to t+1}^{sfg}({{p}_{i}}\in {{\Omega }_{q}})=\begin{cases}
    \Pi (M_{t}^{s}=q)&\text{${r}_{q} \ge {r}_{th}$},\\
    0&\text{${r}_{q} < {r}_{th}$}.
    \end{cases}
\end{equation}
\begin{equation}
    {{r}_{q}}=\frac{\sum\limits_{{{\Omega }_{q}}}{\Pi (M_{t}^{s}=q) \bigodot M_{t\to t+1}^{f}}}{\sum\limits_{{{\Omega }_{q}}}{\Pi (M_{t}^{s}=q)}}
\end{equation}
where $\prod$ is an indicator function and equals 1 if the condition is true, ${{r}_{q}}$ is the ratio to judge the instance is static or dynamic, ${{r}_{th}}=0.5$ is the threshold for judgment.

\subsection{Modified SLAM System with motion removal}
\begin{figure}[thpb]
    \centering
    \includegraphics[width=8cm]{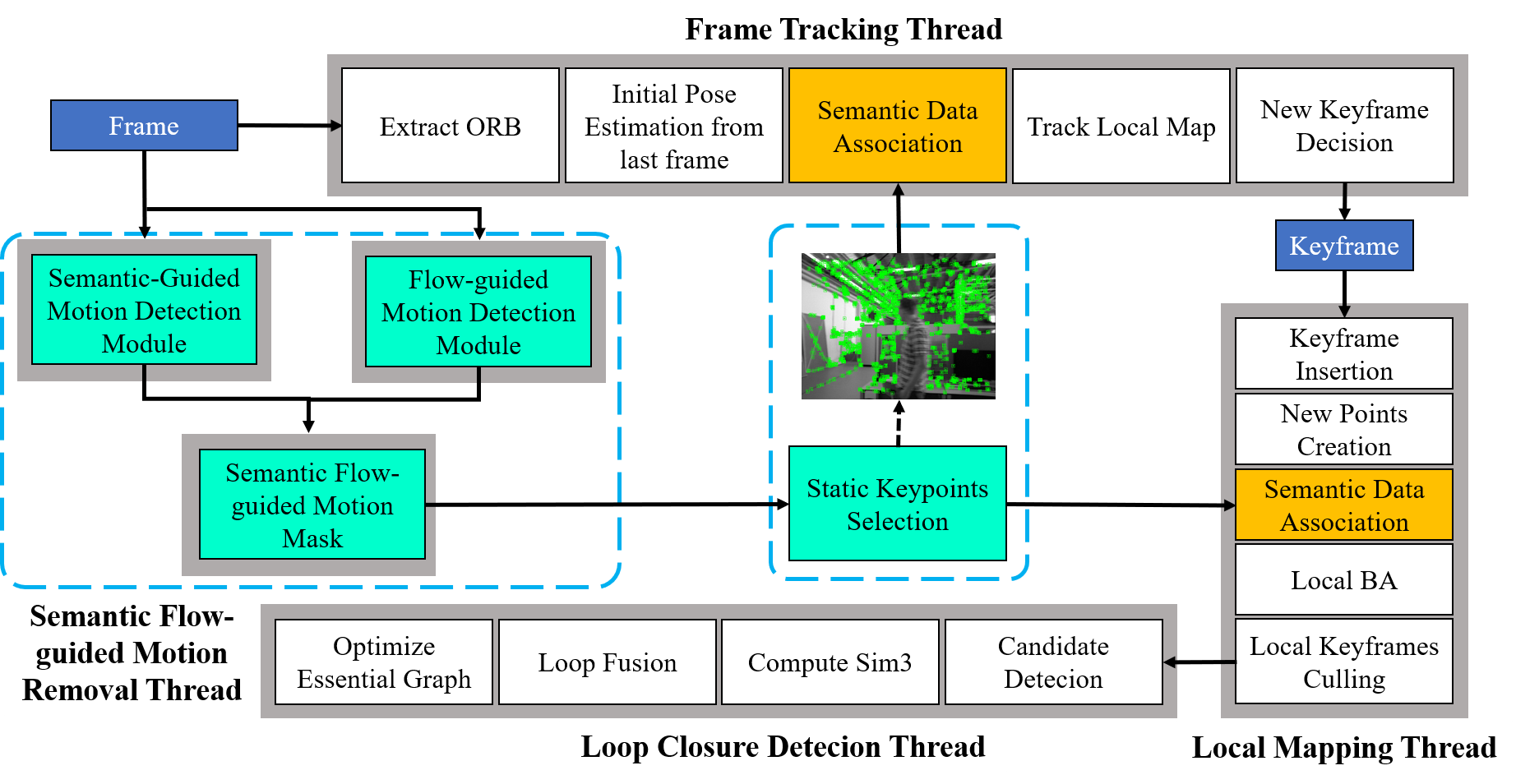}
    \caption{The block diagram of the modified SLAM system with our proposed Semantic Flow-guided motion removal method. We add our Semantic Flow-guided motion removal approach as a separate thread at the front of ORB-SLAM2.}
    \label{block diagram}
\end{figure}

According to the description in \cite{Coarse}, we add the mask $M_{t\to t+1}^{sfg}$ to the frame tracking module and the local mapping module for motion removal. The 2D keypoints of the frame ${{I}_{t}}$ and the corresponding 3D map points with $M_{t\to t+1}^{sfg}=1$ are removed from the Bundle Adjustment (BA) process for camera pose optimization. The modified SLAM system with our proposed Semantic Flow-guided motion removal method is shown in Fig. \ref{block diagram}.

\section{IMPLEMENT DETAILS}

\subsection{Network Architecture}
Our proposed Semantic Flow-guided motion removal framework contains four networks, instance segmentation network, depth prediction network, camera pose prediction network, and optical flow prediction network.  We adopt the Mask R-CNN \cite{maskrcnn} as instance segmentation netwokr ${{\mathcal{N}}_{instance}}$, which is from the Detectron2 system \cite{wu2019detectron2} proposed by Facebook. The model is trained on the COCO dataset \cite{2014Microsoft} and the Cityscape dataset \cite{cordts2016cityscapes} for indoor and outdoor scenes respectively. Then, the model is directly used without any fine-tuning. For the depth prediction network ${{\mathcal{N}}_{depth}}$ and the pose prediction network ${{\mathcal{N}}_{pose}}$, we leverage a state-of-the-art method: Monodepth2 \cite{godard2019digging}.  We use separate ResNet50 \cite{He2016Deep} encoders without shared weight for both the ${{\mathcal{N}}_{depht}}$ and the ${{\mathcal{N}}_{pose}}$. The model is trained on the Depth Eigen Split \cite{eigen2014depth} of the KITTI dataset \cite{2013Vision} and then fine-tuned on the Cityscape dataset \cite{cordts2016cityscapes}. For the optical flow network ${{\mathcal{N}}_{flow}}$, we apply a learning-based dense optical flow estimation network, PWC-Net \cite{sun2018pwc}. The model is trained on the FlyingChairs dataset \cite{mayer2016large} and then fine-tuned on Sintel \cite{butler2012naturalistic} and KITTI training datasets \cite{2013Vision}.

\subsection{Error Metrics}
We use two different metrics, the Absolute Pose Error (APE) proposed in \cite{2012A}, and the Relative Pose Error (RPE) proposed in \cite{2013Vision}. Considering the validity of the metrics, we mainly use the Root Mean Square Error (RMSE) to evaluate the results. The unit of APE and RPE applied in this paper for the TUM dataset and KITTI dataset are shown in Table \ref{unit}.  For metric APE, the translation part stands for the global consistency of trajectory. The metric RPE measures the translational and rotational drift. In the TUM dataset, the errors are measured in cm/frame for translation and degrees per frame for rotation. In the KITTI dataset, our evaluation computes translational and rotational errors for all possible subsequences of length (100, ..., 800) meters. The final value of these errors is according to the average of those values, where errors are measured in percent for translation and degrees per meter for rotation.

\begin{table}[ht]
\caption{The unit of APE and RPE metrics applied in the TUM dataset and KITTI dataset for evaluation}
\label{unit}
\begin{center}
\begin{tabular}{cccc}
\toprule
\multirow{3}{*}{Dataset} & \multicolumn{3}{c}{Unit of the metrics for Evaluation} \\  
                         & APE                 & \multicolumn{2}{c}{RPE}          \\ 
                         & translation   part  & translation part & rotation part \\ 
                        \midrule
TUM                      & cm                  & cm/frame         & deg/frame     \\ 
KITTI                    & m                   & percent          & deg/100m      \\ 
\bottomrule
\end{tabular}
\end{center}
\end{table}

\section{EXPERIMENTS}

\subsection{Indoor evaluation using TUM dynamic dataset}
The TUM RGB-D dataset contains 39 indoor sequences captured by a Microsoft Kinect sensor. In our experiment, a subset of sequences, which belongs to the Dynamic Objects category, was utilized for evaluation. There are two types of sequences in the subset. In the \textit{sitting (s)} sequences, two persons sit at a desk, talk, and gesticulate slightly.  There is a low degree of motion in these sequences. In the \textit{walking (w)} sequences, two persons walk through an office scene, which is highly dynamic and challenging for the SLAM systems. We evaluate our proposed method on this dataset to verify the performance in indoor dynamic scenarios.

\begin{figure}[thpb]
    \centering
        \subfigure[]{\includegraphics[width=2cm]{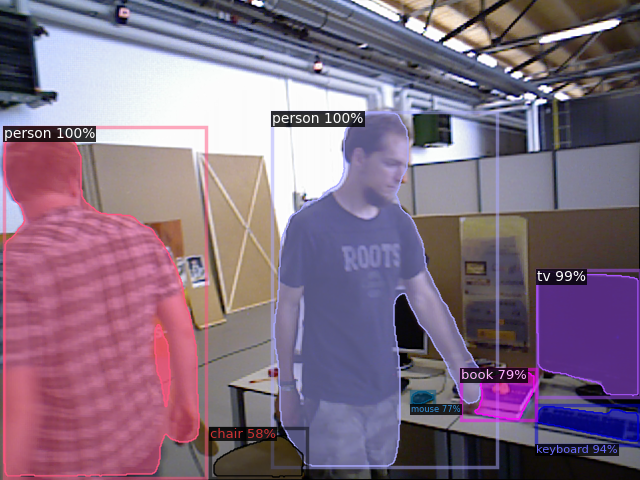}}  
        \subfigure[]{\includegraphics[width=2cm]{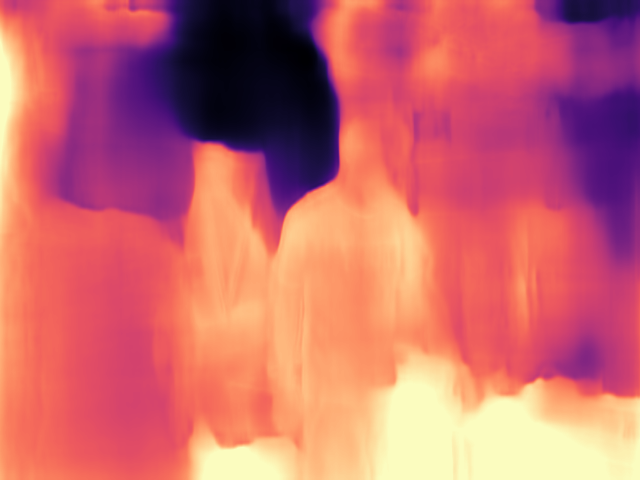}} 
        \subfigure[]{\includegraphics[width=2cm]{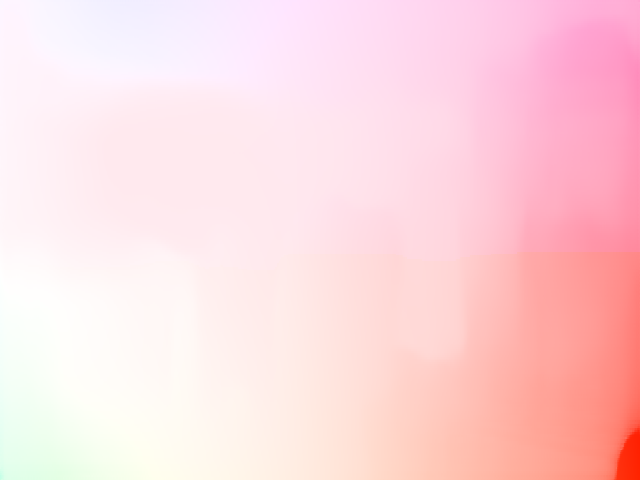}} 
        \subfigure[]{\includegraphics[width=2cm]{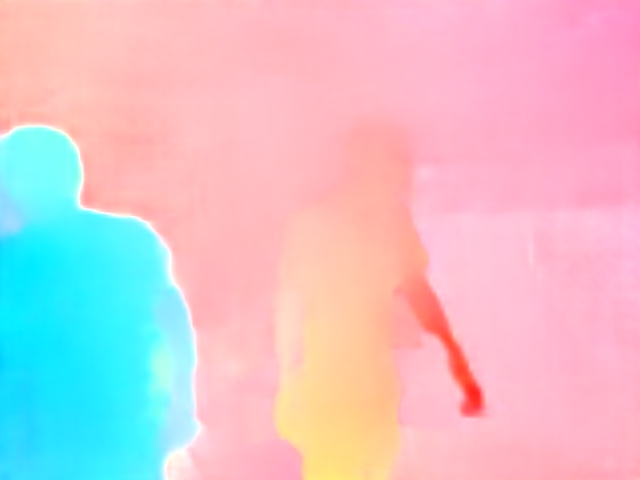}}\\
        \subfigure[]{\includegraphics[width=2cm]{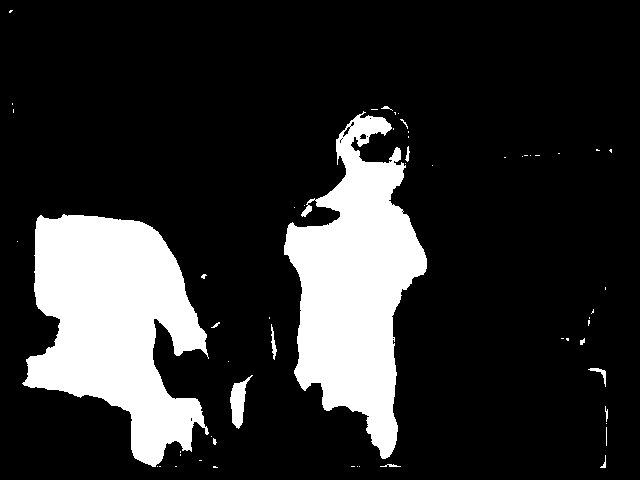}} 
        \subfigure[]{\includegraphics[width=2cm]{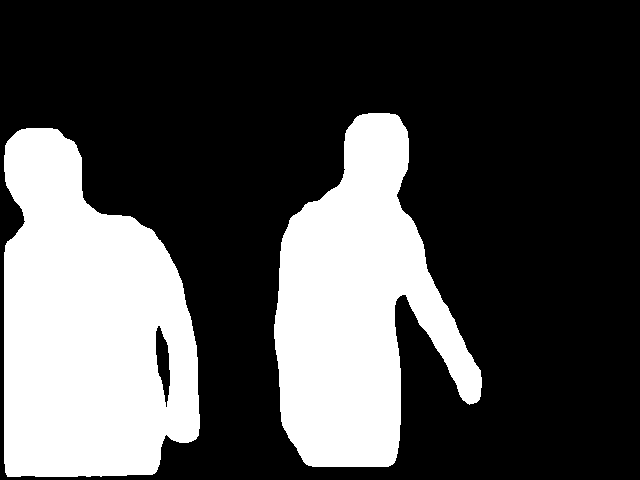}}
        \subfigure[]{\includegraphics[width=2cm]{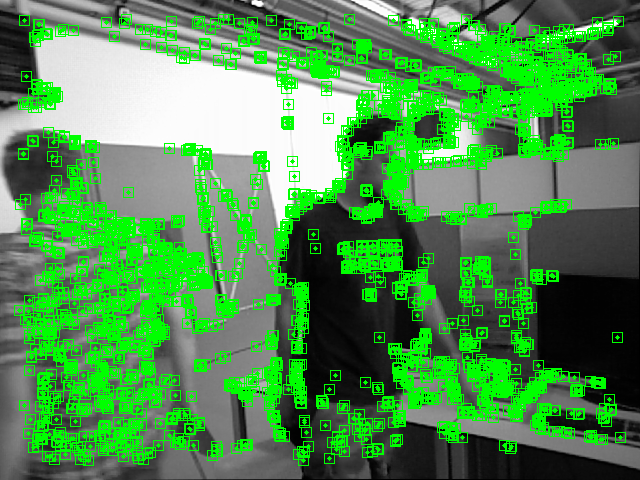}}
        \subfigure[]{\includegraphics[width=2cm]{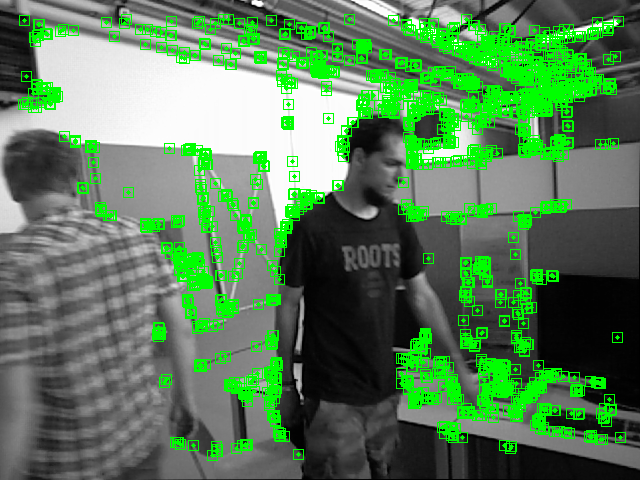}}\\
        
    \caption{The visual results of our proposed Semantic Flow-guided method in the TUM dynamic objects datasets. The first line shows the predictions of four CNNs without any retrain on the validation datasets. (a) shows instance masks. (b) shows depth maps. (c) shows the synthetic rigid flow. (d) shows the predicted optical flow. (e) shows the flow-guided masks. (f) shows the semantic flow-guided masks. (g) and (h) show the ORB keypoints w/o motion removal.}
    \label{tum_show}
\end{figure}

Table \ref{tum_ape} shows the results on the TUM dynamic objects dataset, compared against the original RGB-D ORB-SLAM2 and the Coarse-semantic method. Our proposed method is superior to the RGB-D ORB-SLAM2 in each sequence, especially in high-dynamic \textit{walking (w)} sequences. In these sequences, there is a large deviation in camera pose estimation for RGB-D ORB-SLAM2 due to the existence of moving objects. With our proposed motion removal method, the system can eliminate the influence of moving objects in high-dynamic scenarios (as shown in Fig. \ref{tum_show}). The Coarse-semantic \cite{Coarse} is the system in which the approach removes moving objects rely on category. In comparison, the localization accuracy is significantly lower than the original RGB-D ORB-SLAM2 and Semantic Flow-guided system in low-dynamic scenarios. The fundamental reason is that most of the movable objects in the type of sitting sequences are static. If such objects are eliminated based only on the a priori semantic information, the number of valid keypoints will be greatly reduced for frame tracking and local mapping. Compared with the Coarse-semantic, our proposed method predicts the motion mask by judging each instance's motion state. In this way, we can achieve high camera pose estimation accuracy, whether in high-dynamic or low-dynamic scenarios.

Besides, we also compare our approach with the SOTA SLAM system designed specifically for dynamic environments. Dyna-SLAM \cite{DynaSLAM}, Detect-SLAM \cite{Detectslam}, SOF-SLAM \cite{2019SOF}, DS-SLAM \cite{Qiao2018DS} and Dynamic-SLAM \cite{2019Dynamic} are adopted for comparisons. Table \ref{tum_ape_compare} shows the comparison results between our proposed method and these SOTA SLAM systems mentioned above in the TUM dynamic objects dataset. As we can observe, in low-dynamic scenarios, our Semantic Flow-guided method performs better than all the other methods except the Dynamic-SLAM on the sequence fr3/s/xyz. In high-dynamic scenarios, the accuracy of our system is better than Dyna-SLAM, Detect-SLAM, SOF-SLAM, and DS-SLAM in most sequence. Note that we didn't retrain the networks deliberately on the TUM datasets. The images in the TUM datasets were quite different from the training samples. Thus, the predictions of networks were not accurate but still were usable. Although the error of Dynamic-SLAM is slightly lower than our method on the sequence fr3/w/half and fr3/w/xyz, the difference is acceptable. If the networks are retrained on the TUM datasets, the results on high-dynamic sequences will get better.

\begin{table}[ht]
\caption{Comparison of localization accuracy, Absolute Pose Error (APE) for translation part (Unit: cm) of our method to orb-slam2 (rgb-d) \cite{ORBSLAM2AO} and Coarse-semantic \cite{Coarse} in the TUM dynamic objects dataset}
\label{tum_ape}
\begin{center}
\begin{tabular}{cccc}
\toprule
Sequence             & ORB-SLAM2 \cite{ORBSLAM2AO} & Coarse-Semantic \cite{Coarse} & Ours \\
 \midrule
    fr2/d            & 0.64                       & 0.74                   & \textbf{0.55}        \\
    fr3/s/half       & 2.16                       & 5.33                   & \textbf{1.33}        \\
    fr3/s/rpy        & 2.32                       & 3.76                   & \textbf{1.77}        \\
    fr3/s/static     & 0.90                       & 0.65                   & \textbf{0.63}        \\
    fr3/s/xyz        & 0.92                       & 1.81                   & \textbf{0.89}        \\
    fr3/w/half       & 51.62                      & 3.09                   & \textbf{2.71}        \\
    fr3/w/rpy        & 89.56                      & 2.85                   & \textbf{2.73}        \\
    fr3/w/static     & 39.93                      & \textbf{0.78}          & 0.93                 \\
    fr3/w/xyz        & 55.76                      & \textbf{1.54}          & 1.59 
\\
\bottomrule
\end{tabular}
\end{center}
\end{table}

\begin{table*}[ht]
\caption{Comparison of localization accuracy, Absolute Pose Error (APE) for translation part (Unit: cm) of our method to existing methods in the TUM dynamic objects dataset}
\label{tum_ape_compare}
\begin{center}
\begin{tabular}{ccccccc}
\toprule
Sequence & Dyna-SLAM \cite{DynaSLAM} & Detect-SLAM \cite{Detectslam} & SOF-SLAM \cite{2019SOF} & DS-SLAM \cite{Qiao2018DS} & Dynamic-SLAM \cite{2019Dynamic} & Ours \\
                 \midrule
fr2/desk\_ps     & -                  & -                    & -                 & -              & 1.87                & \textbf{0.55}        \\
fr3/s/half       & 1.70               & 2.31                 & -                 & -              & 1.46                & \textbf{1.33}        \\
fr3/s/rpy        & -                  & -                    & -                 & -              & 3.45                & \textbf{1.77}        \\
fr3/s/static     & -                  & -                    & 1.00              & 0.65           & -                   & \textbf{0.63}        \\
fr3/s/xyz        & 1.50               & 2.01                 & -                 & -              & \textbf{0.60}       & 0.89                 \\
fr3/w/half       & 2.50               & 5.14                 & 2.90              & 3.03           & \textbf{2.14}       & 2.71                 \\
fr3/w/rpy        & 3.50               & 29.59                & \textbf{2.70}     & 44.42          & 6.03                & \textbf{2.73}        \\
fr3/w/static     & \textbf{0.60}      & -                    & 0.70              & 0.81           & -                   & 0.93                 \\
fr3/w/xyz        & 1.50               & 2.41                 & 1.80              & 1.86           & \textbf{1.32}       & 1.59
\\
\bottomrule
\end{tabular}
\end{center}
\end{table*}

\subsection{Outdoor evaluation using KITTI odometry dataset}

\begin{figure}[thpb]
    \centering
        \subfigure[]{\includegraphics[width=4cm]{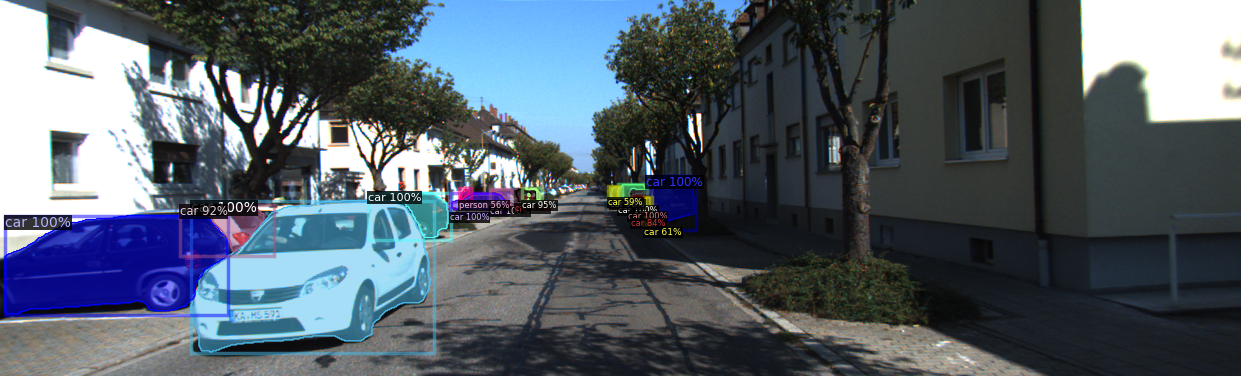}}  
        \subfigure[]{\includegraphics[width=4cm]{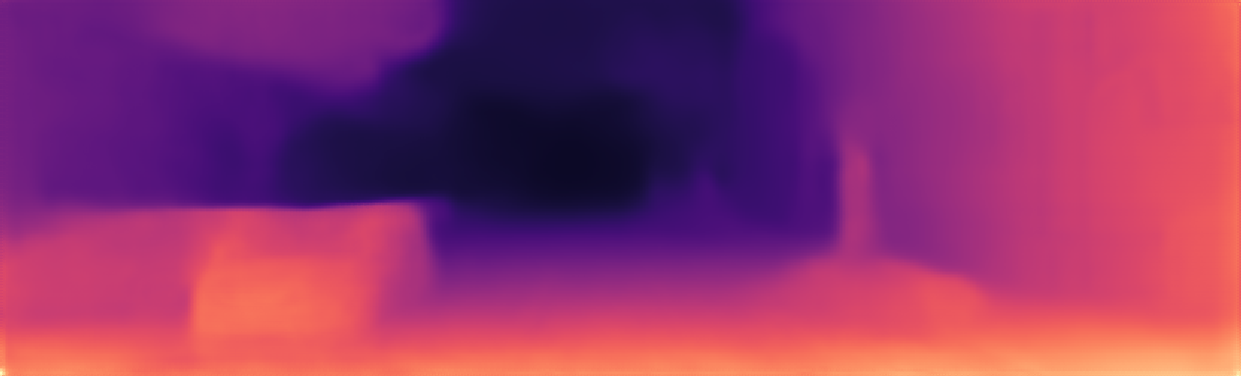}} \\
        \subfigure[]{\includegraphics[width=4cm]{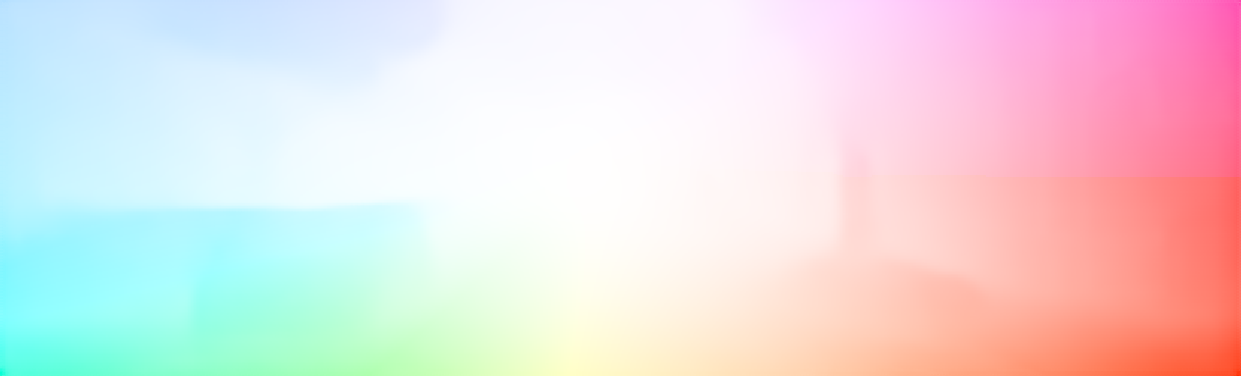}} 
        \subfigure[]{\includegraphics[width=4cm]{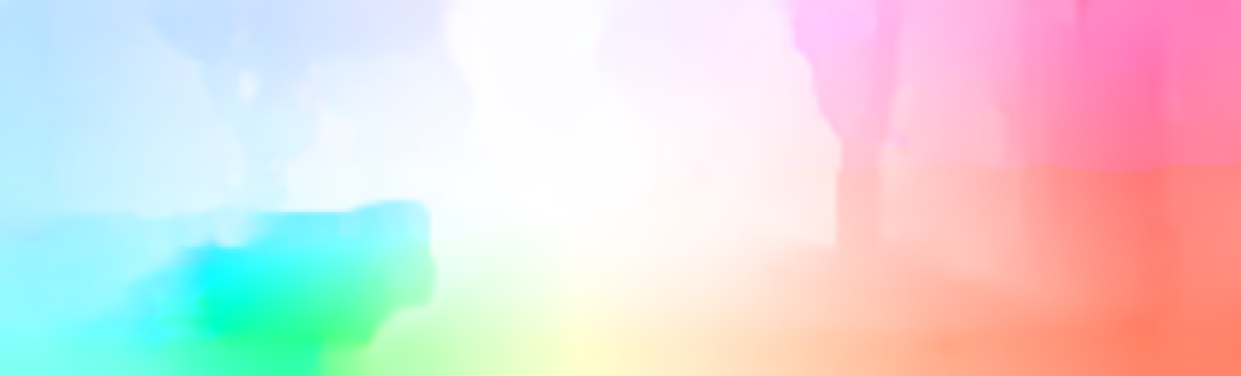}}\\
        \subfigure[]{\includegraphics[width=4cm]{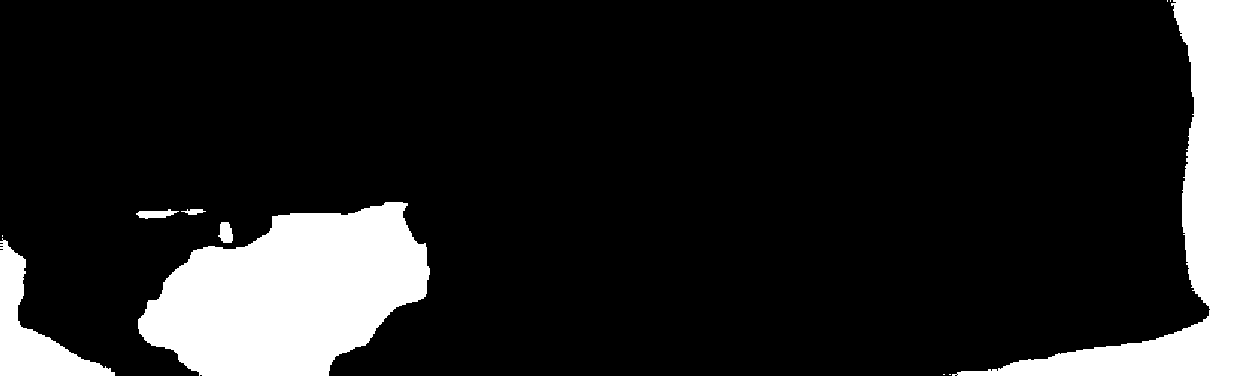}} 
        \subfigure[]{\includegraphics[width=4cm]{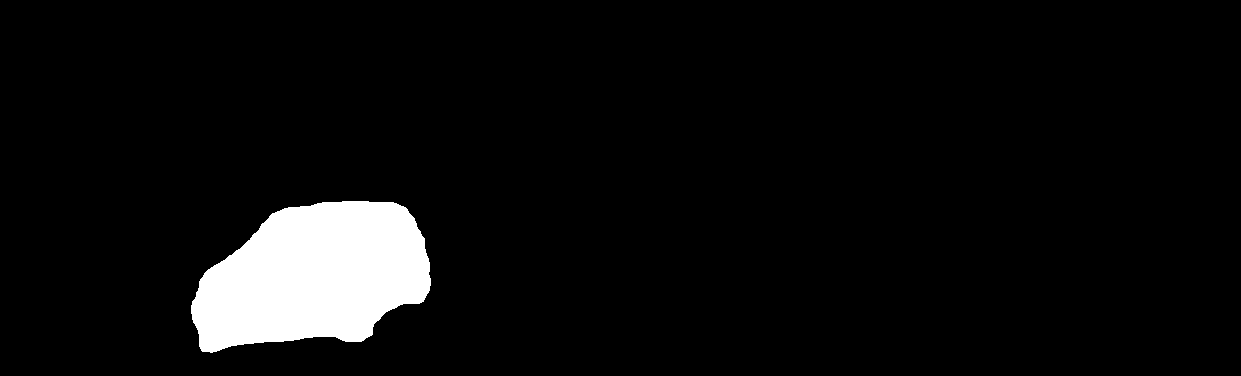}}\\
        \subfigure[]{\includegraphics[width=4cm]{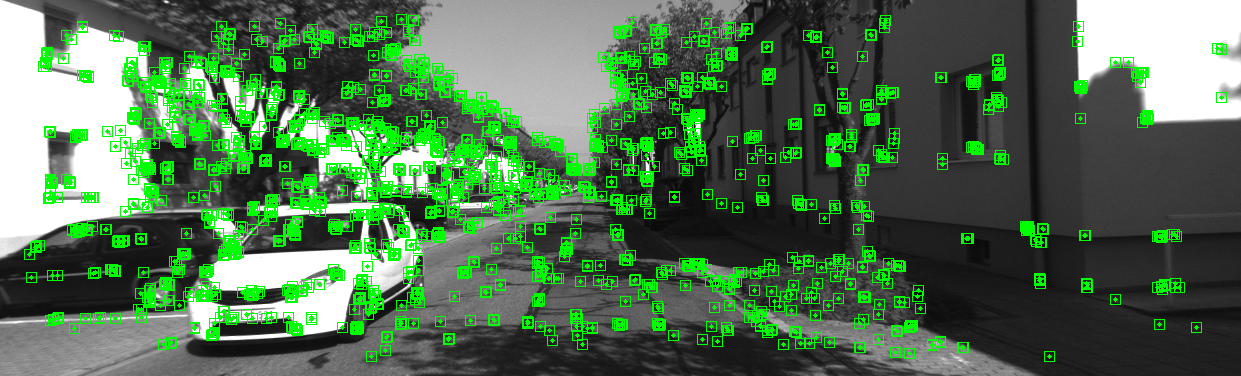}}
        \subfigure[]{\includegraphics[width=4cm]{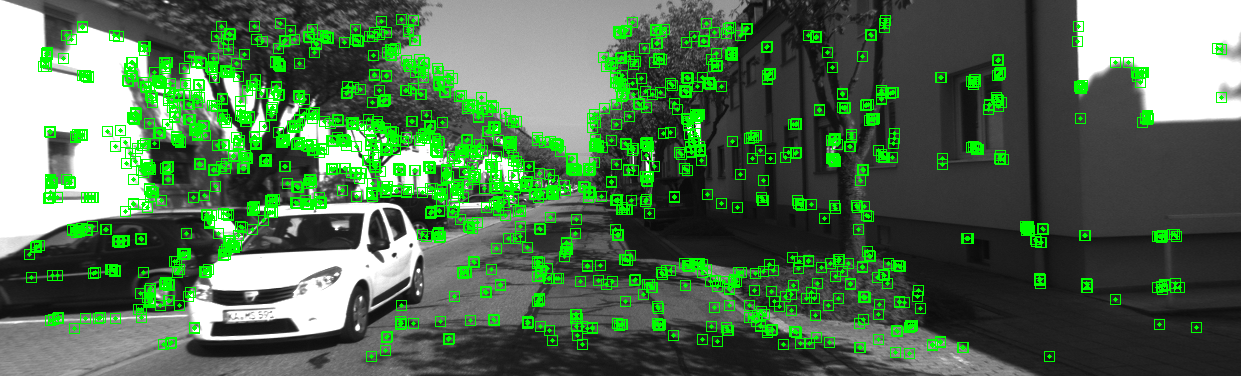}}\\
        
    \caption{The example of our proposed semantic flow-guided motion removal method in the KITTI odometry dataset. (a) shows instance masks. (b) shows depth maps. (c) shows the synthetic rigid flow. (d) shows the predicted optical flow. (e) shows the flow-guided masks. (f) shows the semantic flow-guided masks. (g) and (h) show the ORB keypoints w/o motion removal.}
    \label{kitti_show}
\end{figure}

\begin{table*}[ht]
\caption{Comparison of localization accuracy of our method to existing methods in the KITTI odometry dataset}
\label{kitti_compare}
\begin{center}
\begin{tabular}{ccccccccccccc}
\toprule
\multirow{2}{*}{Sequence} & \multicolumn{3}{c}{ORB-SLAM2 (Stereo) \cite{ORBSLAM2AO}}                                                                                                                                                 & \multicolumn{3}{c}{DynaSLAM \cite{DynaSLAM}}                                                                                                                                                             & \multicolumn{3}{c}{Coarse-semantic \cite{Coarse}}                                                                                                                                                      & \multicolumn{3}{c}{Ours}                                                                                                                                                        \\
                          & \begin{tabular}[c]{@{}c@{}}${{t}_{rel}}$\\ (\%)\end{tabular} & \begin{tabular}[c]{@{}c@{}}${{r}_{rel}}$\\ (deg/100m)\end{tabular} & \begin{tabular}[c]{@{}c@{}}${{t}_{abs}}$\\ (m)\end{tabular} & \begin{tabular}[c]{@{}c@{}}${{t}_{rel}}$\\ (\%)\end{tabular} & \begin{tabular}[c]{@{}c@{}}${{r}_{rel}}$\\ (deg/100m)\end{tabular} & \begin{tabular}[c]{@{}c@{}}${{t}_{abs}}$\\ (m)\end{tabular} & \begin{tabular}[c]{@{}c@{}}${{t}_{rel}}$\\ (\%)\end{tabular} & \begin{tabular}[c]{@{}c@{}}${{r}_{rel}}$\\ (deg/100m)\end{tabular} & \begin{tabular}[c]{@{}c@{}}${{t}_{abs}}$\\ (m)\end{tabular} & \begin{tabular}[c]{@{}c@{}}${{t}_{rel}}$\\ (\%)\end{tabular} & \begin{tabular}[c]{@{}c@{}}${{r}_{rel}}$\\ (deg/100m)\end{tabular} & \begin{tabular}[c]{@{}c@{}}${{t}_{abs}}$\\ (m)\end{tabular} \\
                          \midrule
00                        & 0.70                 & 0.25                       & 1.30                & 0.74                 & 0.26                       & 1.40                & 1.48                 & 0.36                       & 4.44                & \textbf{0.68}        & \textbf{0.24}              & \textbf{1.23}       \\
01                        & 1.39                 & 0.21                       & 10.40               & 1.57                 & 0.22                       & 9.40                & 1.25                 & 0.21                       & 9.41                & \textbf{1.38}        & \textbf{0.16}              & \textbf{9.07}       \\
02                        & 0.76                 & \textbf{0.23}              & 5.70                & 0.80                 & 0.24                       & 6.70                & 1.56                 & 0.31                       & 8.43                & \textbf{0.74}        & \textbf{0.23}              & \textbf{4.99}       \\
03                        & 0.71                 & 0.18                       & 0.60                & 0.69                 & 0.18                       & 0.60                & 2.70                 & 0.18                       & 4.64                & \textbf{0.66}        & \textbf{0.17}              & \textbf{0.55}       \\
04                        & 0.48                 & 0.13                       & 0.20                & 0.45                 & 0.09                       & 0.20                & 0.94                 & 0.13                       & 0.99                & \textbf{0.45}        & \textbf{0.08}              & \textbf{0.17}       \\
05                        & 0.40                 & \textbf{0.16}              & 0.80                & 0.40                 & \textbf{0.16}              & 0.80                & 0.65                 & 0.21                       & 1.72                & \textbf{0.37}        & \textbf{0.16}              & \textbf{0.72}       \\
06                        & 0.51                 & 0.15                       & 0.80                & 0.50                 & 0.17                       & 0.80                & 0.76                 & 0.24                       & 1.80                & \textbf{0.43}        & \textbf{0.14}              & \textbf{0.70}       \\
07                        & 0.50                 & 0.28                       & 0.50                & 0.52                 & 0.29                       & 0.50                & 0.68                 & 0.26                       & 1.13                & \textbf{0.38}        & \textbf{0.26}              & \textbf{0.44}       \\
08                        & 1.05                 & 0.32                       & 3.60                & 1.05                 & 0.32                       & 3.50                & 1.19                 & 0.33                       & 3.93                & \textbf{1.00}        & \textbf{0.31}              & \textbf{3.32}       \\
09                        & 0.87                 & 0.27                       & 3.20                & 0.93                 & 0.29                       & 1.60                & 1.18                 & 0.26                       & 4.68                & \textbf{0.81}        & \textbf{0.22}              & \textbf{1.57}       \\
10                        & 0.60                 & 0.27                       & 1.00                & 0.67                 & 0.32                       & 1.20                & 0.98                 & 0.27                       & 2.45                & \textbf{0.57}        & \textbf{0.24}              & \textbf{0.97}   
\\
\bottomrule
\end{tabular}
\end{center}
\end{table*}

The KITTI odometry dataset \cite{2013Vision} contains stereo sequences recorded from a moving vehicle in urban and highway environments. The stereo camera has a ~54cm baseline and works at 10Hz.  In urban’s sequences, most vehicles are stationary and parked at the side of the road, including a small number of moving vehicles, bicycles, motorcycles, and pedestrians. In the highway sequences (seq 01, 12, for instance), most vehicles are moving. We evaluate our proposed method on this dataset to verify the performance in outdoor dynamic environments.

We adopt the evaluation metrics mentioned in \cite{ORBSLAM2AO} which is described in Section IV.B.. We use ${{t}_{abs}}$ to denote the translation part of APE, ${{t}_{rel}}$ and ${{r}_{rel}}$ to denote the translation part and rotation part of RPE. Note that the comparison results in Table \ref{kitti_compare} also shows the superior performance of our method.  DynaSLAM detects the moving objects combining both the geometrical and semantical approaches. This method can improve the accuracy of pose estimation in relative high dynamic scenarios (seq 01) but does not perform well in some low-dynamic sequences, such as seq 00, 02, 10. The reason is the same as in the Coarse-semantic method that it can not judge the motion state of the movable objects. Compared with the Coarse-semantic and DynaSLAM system, our Semantic Flow-guided method provide a motion mask by determining the motion state of each instance. Therefore, whether in static or dynamic outdoor scenarios, our proposed motion removal method can obtain accurate and stable camera trajectories.
Fig. \ref{kitti_show} shows the visual results of our proposed method on the KITTI odometry dataset. The example of instance segmentation maks shown in Fig. \ref{kitti_show}(a) includes a moving vehicle and stationary vehicles parked at the roadside. The flow-guided mask Fig. \ref{kitti_show}(e) can be directly used to detect moving vehicles, but the mask is not accurate enough to cover the whole instance similar to the instance mask. With the combination of these two masks, only the dynamic objects are reserved in the semantic flow-guided mask Fig. \ref{kitti_show}(f). Therefore, the keypoints belongs to moving objects are removed while the keypoints on static objects are reserved, as shown in Fig. \ref{kitti_show}(h).

\subsection{Comparison for several variants of semantic flow-guided motion removal method}

\begin{table}[ht]
\caption{Absolute Pose Error (APE) for translation part (RMSE Unit: cm) for several variants of our approach in the TUM dynamic objects dataset}
\label{tum_ape_self_compare}
\begin{center}
\begin{tabular}{cccc}
\toprule
Sequence     & Semantic-Guided & Flow-Guided  & Semantic Flow-Guided \\
\midrule
    fr2/d        & 0.61                 & 0.56             & \textbf{0.55}              \\
    fr3/s/half   & 1.64                 & \textbf{1.33}    & \textbf{1.33}              \\
    fr3/s/rpy    & 2.40                 & 1.79             & \textbf{1.77}              \\
    fr3/s/static & \textbf{0.56}        & 0.67             & 0.63                       \\
    fr3/s/xyz    & 1.22                 & 0.92             & \textbf{0.89}              \\
    fr3/w/half   & \textbf{2.34}        & 12.85            & 2.71                       \\
    fr3/w/rpy    & 2.87                 & 2.85             & \textbf{2.73}              \\
    fr3/w/static & \textbf{0.73}        & 0.98             & 0.93                       \\
    fr3/w/xyz    & \textbf{1.50}        & 1.64             & 1.59  
\\
\bottomrule
\end{tabular}
\end{center}
\end{table}

\begin{table}[ht]
\caption{Absolute Pose Error (APE) for translation part (RMSE Unit: cm) for several variants of our approach in the KITTI odometry dataset}
\label{kitti_ape_self_compare}
\begin{center}
\begin{tabular}{cccc}
\toprule
Sequence & Semantic-Guided & Flow-Guided & Semantic Flow-Guided \\
\midrule
    00               & 1.28                 & 1.38             & \textbf{1.23}              \\
    01               & 11.25                & 9.31             & \textbf{9.07}              \\
    02               & 5.45                 & 5.50             & \textbf{4.99}              \\
    03               & 0.63                 & 0.65             & \textbf{0.55}              \\
    04               & 0.23                 & 0.18             & \textbf{0.17}              \\
    05               & 0.77                 & 0.76             & \textbf{0.72}              \\
    06               & 0.75                 & 0.82             & \textbf{0.70}              \\
    07               & 0.53                 & 0.54             & \textbf{0.44}              \\
    08               & 3.57                 & 3.62             & \textbf{3.32}              \\
    09               & 1.67                 & 2.62             & \textbf{1.57}              \\
    10               & 1.10                 & 0.93             & \textbf{0.97}
\\
\bottomrule
\end{tabular}
\end{center}
\end{table}

The results of different variations of our system are shown in Table \ref{tum_ape_self_compare} and Table \ref{kitti_ape_self_compare}. Semantic-guided is the system in which only the semantic-guided module is applied to segment out the a priori dynamic objects. For Flow-guided, the moving objects are detected by the flow-guided module that incorporates the rigid flow and optical flow into the system. Semantic Flow-Guided stands for the system that predicts the final motion mask by combining both the semantic and flow information. According to Table 5, in low-dynamic scenarios (sitting) of the TUM dataset, the system using semantic flow-guided masks is the most accurate one in most sequences. The system using semantic-guided masks has a higher error due to most movable objects in these sequences are static. In high-dynamic scenarios (walking) of the TUM dataset, the error of system using semantic flow-guided masks is slightly higher than the one using semantic-guided masks in most sequences. 

\section{CONCLUSIONS}
For indoor and outdoor environments, moving objects will lead to the SLAM systems' low accuracy during mapping. Thus, motion removal can increase the precision of location and mapping. This paper presents an approach, named Semantic Flow-guided, that integrates semantic information and optical flow information to segment motion regions. In Semantic Flow-guided, we used four CNNs to predict depth, pose, optical flow, and instance semantic mask for each frame in parallel. Utilize rigid flow, synthesized by the depth and pose, and predicted optical flow to get the initial motion masks. We then proposed a finetune method based on k-means to employ instance semantic masks to get semantic flow-guided masks. After finetuning, semantic flow-guided masks can be adopted to remove moving objects. We evaluated our approach on the TUM datasets and KITTI odometry datasets. The experimental results show that Semantic Flow-guided method could significantly improve the performance of ORB-SLAM2 in dynamic scenes, both in indoor and outdoor. Compared with previous works, our method outperforms them both in low-dynamic and high-dynamic sequences.

In our future works, we will try to add the Semantic Flow-guided mask to our learning-based visual odometry to improve the deep learning-based SLAM system's performance. We believe that our motion removal method also can be used in the learning-based loop closure detection method \cite{2020Compressed} to remove moving objects in feature maps.

\addtolength{\textheight}{-12cm}   








\bibliographystyle{unsrt}
\bibliography{references}
\end{document}